\pdfoutput=1
\documentclass{article}

\usepackage[preprint]{neurips_2024}

\usepackage[utf8]{inputenc} 
\usepackage[T1]{fontenc}    
\usepackage{hyperref}       
\usepackage{url}            
\usepackage{booktabs}       
\usepackage{amsfonts}       
\usepackage{nicefrac}       
\usepackage{microtype}      
\usepackage{xcolor}         
\usepackage{graphicx}
\usepackage{amsmath}
\usepackage{cleveref}
\usepackage{geometry}
\usepackage{algorithm}
\usepackage{algorithmic}
\usepackage{multirow} 
\usepackage{wrapfig}
\usepackage[numbers]{natbib}
\hypersetup{
hidelinks,
colorlinks=true,
}
\newtheorem{theorem}{Theorem}

\title{ExactDreamer: High-Fidelity Text-to-3D Content Creation via Exact Score Matching}

\author{
    \textbf{Yumin Zhang}\textsuperscript{1}\thanks{Equal contribution}\quad 
    \textbf{Xingyu Miao}\textsuperscript{2}\footnotemark[1]\quad 
   \textbf{Haoran Duan}\textsuperscript{1}\footnotemark[1]\quad \\
   \textbf{Bo Wei}\textsuperscript{1}\quad 
    \textbf{Tejal Shah}\textsuperscript{1}\quad 
    \textbf{Yang Long}\textsuperscript{2}\quad 
    \textbf{Rajiv Ranjan}\textsuperscript{1}\quad \\
\\
\small
    \textsuperscript{1}Newcastle University, UK\quad \textsuperscript{2}Durham University, UK \quad 
}

\begin{document}

\maketitle

\begin{abstract}
Text-to-3D content creation is a rapidly evolving research area. Given the scarcity of 3D data, current approaches often adapt pre-trained 2D diffusion models for 3D synthesis. 
Among these approaches, Score Distillation Sampling (SDS) has been widely adopted. 
However, the issue of over-smoothing poses a significant limitation on the high-fidelity generation of 3D models.
To address this challenge, LucidDreamer replaces the Denoising Diffusion Probabilistic Model (DDPM) in SDS with the Denoising Diffusion Implicit Model (DDIM) to construct Interval Score Matching (ISM). However, ISM inevitably inherits inconsistencies from DDIM, causing reconstruction errors during the DDIM inversion process. This results in poor performance in the detailed generation of 3D objects and loss of content.
To alleviate these problems, we propose a novel method named Exact Score Matching (ESM). Specifically, ESM leverages auxiliary variables to mathematically guarantee exact recovery in the DDIM reverse process. Furthermore, to effectively capture the dynamic changes of the original and auxiliary variables, the LoRA of a pre-trained diffusion model implements these exact paths.
Extensive experiments demonstrate the effectiveness of ESM in text-to-3D generation, particularly highlighting its superiority in detailed generation.
Code is available at: \href{https://github.com/zymvszym/ExactDreamer}{https://github.com/zymvszym/ExactDreamer}.
\end{abstract}

\section{Introduction}
\begin{figure}[ht]
    \centering    \includegraphics[width=1.0\linewidth]{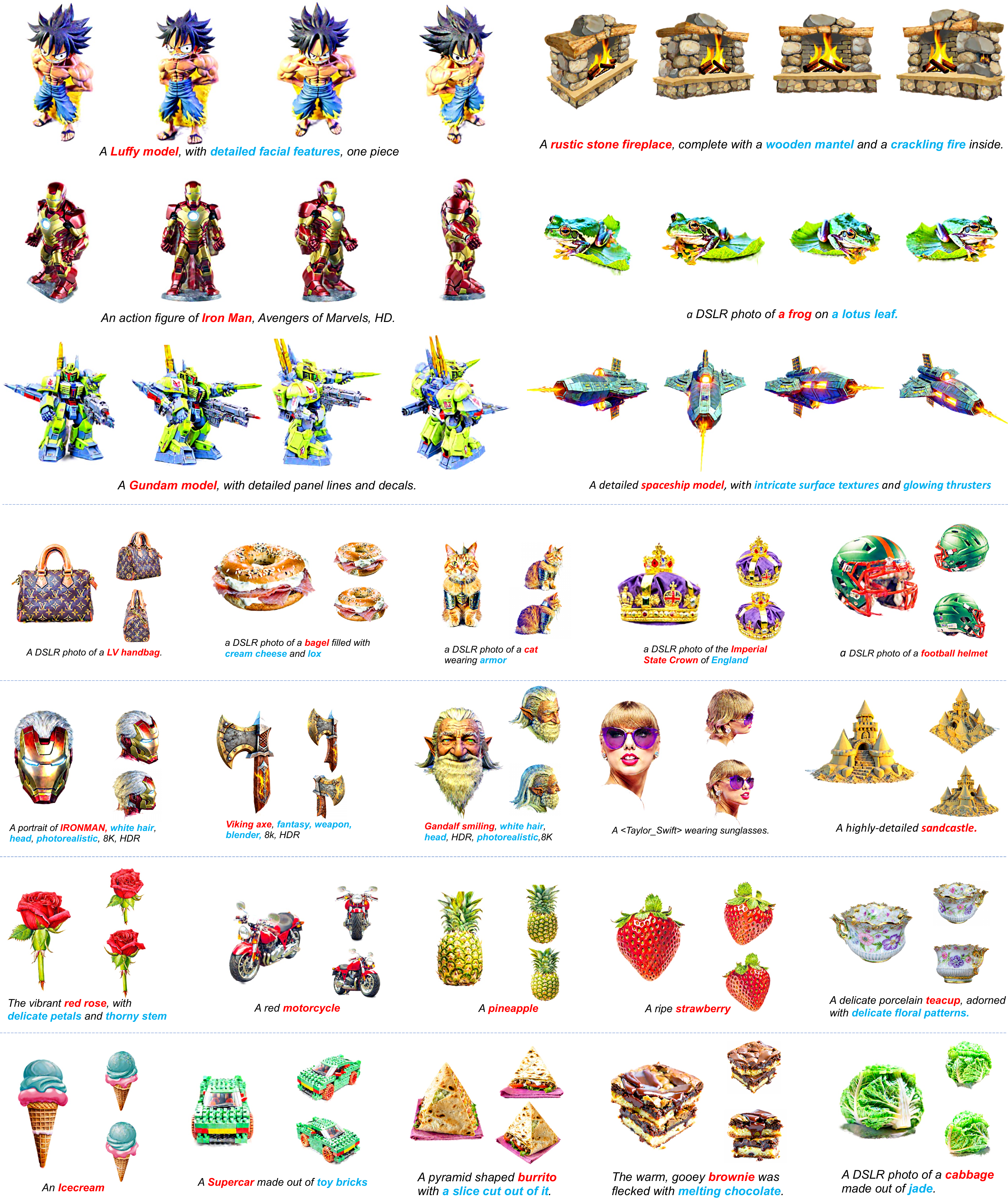}
    \caption{\textbf{Text-to-3D samples generated from our framework.}
    We propose a novel method named Exact Score Matching (ESM) that utilizes the pre-trained 2D diffusion model to guide high-fidelity content generation.
    The generative 3D results illustrate the superiority of ESM.}
    \label{fig:fig1}
\end{figure}

3D content creation plays a crucial role across multiple domains such as virtual and augmented reality, game design, head modeling, and more.
Among these, text-to-3D generation allows for the synthesis of imaginative 3D models guided by text captions and has become a burgeoning research area in recent years.
Despite the significant advancements brought about by deep learning techniques, previous methods~\cite{chen2019text2shape, wu2016learning, duan2023dynamic,duan2020sofa} are limited by the scarcity of 3D data. 

Benefiting the advantages of diffusion models~\cite{ho2020denoising, song2020score}, which have achieved remarkable progress in Text-to-Image (T2I)~\cite{ruiz2023dreambooth,saharia2022photorealistic}, recent works attempt to incorporate pre-trained T2I diffusion models into the text-to-3D generation task, aiming to synthesize high-fidelity 3D models without relying on extensive 3D datasets~\cite{lin2023magic3d,raj2023dreambooth3d}.
In this research area, Score Distillation Sampling (SDS) proposed in DreamFusion~\cite{poole2022dreamfusion} has been widely adopted due to its impressive generation quality.

However, as indicated in~\cite{liang2023luciddreamer, wang2024prolificdreamer}, issues such as over-smoothing hinder the generation of detailed, high-fidelity 3D models.
Various works have been proposed to address the limitations of SDS. 
For instance, ProlificDreamer~\cite{wang2024prolificdreamer} treats the modeled 3D parameters as random variables and thus proposes Variational Score Distillation (VSD), achieving a better balance between high-fidelity and diversity results.
However, the long training time required for VSD is a significant drawback. 
From another perspective, LucidDreamer~\cite{liang2023luciddreamer} introduces Interval Score Matching (ISM), employing Denoising Diffusion Implicit Models (DDIM) ~\cite{song2020denoising} to mitigate the over-smoothing caused by the averaging effect during the 3D generation process, and the experimental results demonstrate ISM can synthesis competitive 3D models within a short time.
Nevertheless, as mentioned in~\cite{hertz2022prompt}, DDIM is unstable in many cases.
The encoding process from realistic samples to noise vectors and the reverse often results in inexact reconstructions, leading to inconsistent generation. 

Considering the realistic requirement of generating 3D models within a limited time, we propose a novel method named Exact Score Matching (ESM) based on ISM, which focuses on alleviating the inconsistencies in the DDIM process and enhancing high-fidelity content generation. 
Specifically, we first illustrate how the inherent bias in ISM can be alleviated by releasing the assumption of local linear approximation. 
Inspired by recent research~\cite{wallace2023edict}, we introduce auxiliary noise variables that facilitate exact recovery during DDIM processes. 
To effectively merge these auxiliary noise variables with the original noise variables, we employ an interactive recovery path. 
This path is implemented by the Low-rank Adaptation (LoRA)~\cite{hu2021lora, ryu2023low} of a pre-trained 2D diffusion model, which also serves to adaptively capture the dynamic changes of these integrated variables.
Overall, our main contributions can be summarized as follows:
\begin{itemize}
    \item We analyze the local linear approximation in DDIM, which hinders the detailed 3D model generation.
    Based on this, we propose a novel text-to-3D method named ESM, aiming to construct an exact recovery and thus enhance the consistencies of 3D model generation. 
    
    \item By utilizing auxiliary noise variables, ESM constructs an exact recovery strategy. 
    Besides, the LoRA is leveraged to capture the dynamic change of the original and auxiliary noise variables effectively and adaptatively. 
    
    \item We conduct extensive experiments, and the results demonstrate the effectiveness of ESM, particularly in generating high-fidelity and detailed models.
\end{itemize}

\section{Related Work}
\subsection{Diffusion models}
Diffusion models~\cite{ho2020denoising,song2020score} have become the dominant method in image generation and have shown potential for application in 3D generation.
Generally, based on the types of 3D data, diffusion models have been widely applied across various 3D representations, such as NeRF~\cite{mildenhall2021nerf}, Gaussian Splatting~\cite{kerbl20233d}, point clouds, and others.
For example, \cite{luo2021diffusion, zhou20213d} generate the point clouds using diffusion algorithms. 
Point-E~\cite{nichol2022point} first uses a T2I diffusion model to generate images, which are then used as conditions for the second step of 3D point cloud synthesis.
Additionally, Shape-E~\cite{jun2023shap} is proposed to generate the NeRF representation of 3D objects, and the Gaussian splatting is adapted as a generative setting in DreamGaussian~\cite{tang2023dreamgaussian}.
Due to the impressive performance of Gaussian splatting on 3D models, our work adopts it as the 3D learnable representation.

\subsection{Text-to-Image Generation}
Impressive achievements have been made in the field of T2I generation.
Early works can be mainly grouped into GAN-based~\cite{reed2016generative, zhang2017stackgan, xu2018attngan, li2019controllable} and autoregressive-based~\cite{ramesh2021zero, ding2021cogview, wu2022nuwa} methods, which are limited to small-scale datasets and sequential error accumulation, respectively.
More recently, methods~\cite{nichol2021glide,rombach2022high, saharia2022photorealistic} integrating diffusion models into T2I have achieved state-of-the-art performance.
Among them, the first work GLIDE~\cite{nichol2021glide} replaces the original class label with text, thus achieving text-guided generation.
Following GLIDE, Imagen~\cite{saharia2022photorealistic} adopts a pre-trained language model as the text encoder.
Moreover, diffusion-based methods are also emerging in some highly relevant areas, such as text-guided art generation~\cite{rombach2022text, jain2023vectorfusion,huang2022draw} and text-guided image editing~\cite{kim2022diffusionclip, kwon2022diffusion}.

\subsection{Text-to-3D Generation}
To generate the 3D content guided by the text prompts,  current text-to-3D methods typically utilize differentiable rendering techniques to obtain images aligned with the text.
For example, DreamField~\cite{jain2022zero} combines NeRF~\cite{mildenhall2021nerf} with a pre-trained CLIP~\cite{radford2021learning} model to achieve text-to-3D generation.
DreamFusion~\cite{poole2022dreamfusion} proposes the score distillation sampling (SDS) loss with volumetric representations used in NeRF to generate high-fidelity 3D content.
Further, based on the SDS, Magic3D~\cite{lin2023magic3d} introduces a coarse-to-fine optimization approach, starting with a coarse representation and then refining the 3D mesh model using differentiable rendering and SDS. 
In a similar two-stage manner, Fantasia3D~\cite{chen2023fantasia3d} utilizes DMTet~\cite{shen2021deep} to obtain geometric knowledge and proposes a bidirectional reflectance distribution function to model the appearance.
Besides, aiming to address the over-smoothing in SDS-based methods, ProlificDreamer~\cite{wang2024prolificdreamer} introduces variational score distillation (VSD) treating the input prompt as a random variable rather than a constant. 
LucidDreamer combines interval score matching (ISM) with DDIM~\cite{song2020denoising} to mitigate over-smoothing caused by naive averaging.
However, the local linear approximation in the DDIM inversion process results in instability during the generation in many cases~\cite{hertz2022prompt}.
To address the inconsistencies caused by this approximation, we introduce auxiliary variables to construct an exact recovery path.

\section{Methodology}
\begin{figure}[!t]
    \centering
    \includegraphics[width=1.0\linewidth]{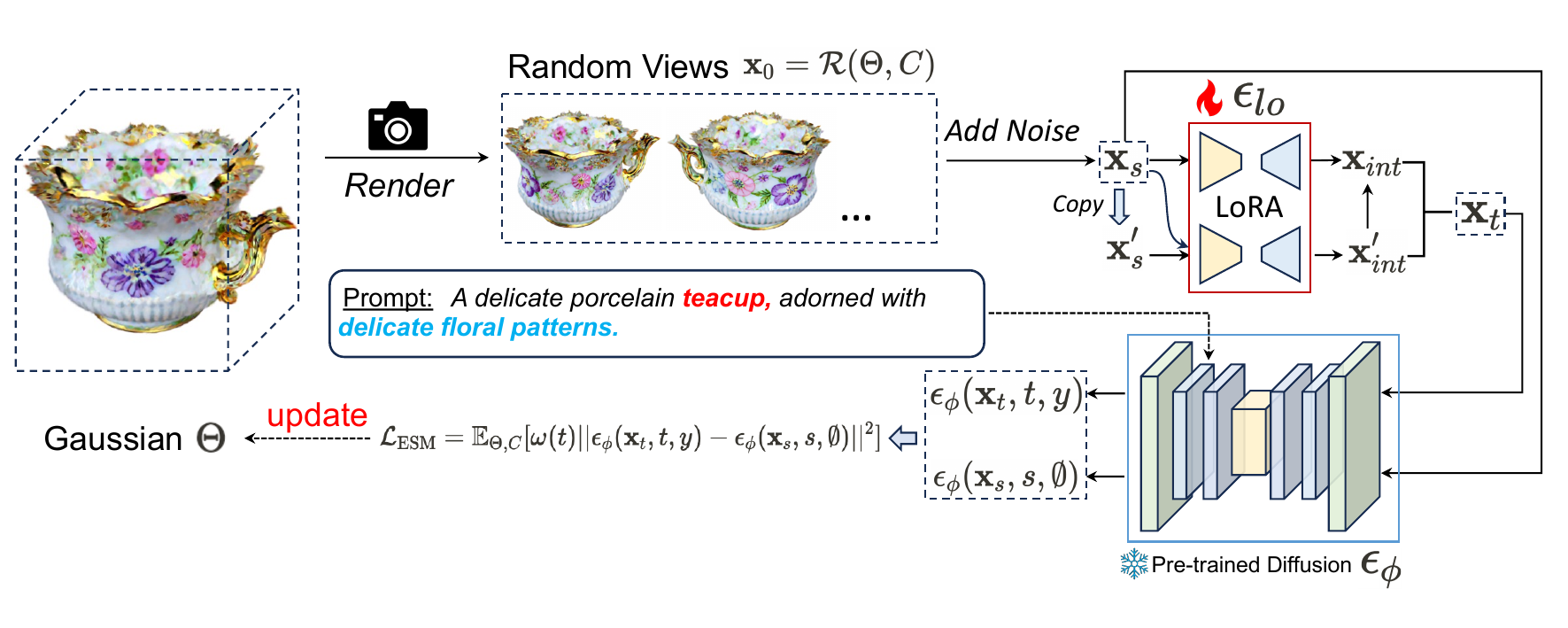}
    \caption{\textbf{Overview of our framework}. 
    Under the camera pose $C$, the Gaussian Splatting $\Theta$ is rendered to 2D image $\mathbf{x}_{0} = \mathcal{R}(\Theta, C)$ and then approach to $\mathbf{x}_{s}$ via DDIM inversion.
    To construct the exact recovery path, we introduce an auxiliary variable $\mathbf{x}_{s}'$ that is copied from $\mathbf{x}_{s}$.
    Intermediate variables $\mathbf{x}_{t}$ and $\mathbf{x}_{t}'$ are estimated by the LoRA, and then mixed to obtain $\mathbf{x}_{t}$.
    Finally, $\Theta$ is updated via optimizing $\mathcal{L}_{\rm{ESM}}$ calculated by $\mathbf{x}_{s}$ and $\mathbf{x}_{t}$.
    }
    \label{fig:overview}
\end{figure}
In this section, we first review the preliminaries of 3D Learnable Representations, the DDIM inversion process, and ISM (see Section~\ref{bg}), which are foundational for our exploration. 
We then introduce our proposed method, Exact Score Matching (ESM), based on ISM.
Specifically, our main contribution is addressing the inherent bias in DDIM by introducing auxiliary variables to design an exact inversion process. 
Considering that the noise latent changes during the optimization process, we utilize LoRA to adaptively model these dynamic changes (see Section~\ref{ESM}). Additionally, we analyze the differences between ISM and ESM (see Section~\ref{com}).
The overview of our method is shown in Figure~\ref{fig:overview}.
\subsection{Preliminaries}
\label{bg}

\subsubsection{3D Learnable Representations}
Existing text-to-3D generation works that primarily incorporate NeRF~\cite{mildenhall2021nerf} have shown that rendering resolution and batch size settings during the training process significantly influence the quality of the final 3D creation. 
Considering the substantial computational resources required by increasing rendering resolution, we adopt 3D Gaussian Splatting~\cite{kerbl20233d}, an efficient explicit rendering method, for both rendering and optimization as the 3D representation.
Specifically, given a learnable 3D representation $\Theta$, the differentiable rendering function $\mathcal{R}(\Theta, C)$ can be defined to render an image under the camera pose $C$.

\subsubsection{Review of DDIM inversion}
The DDIM inversion sampling scheme proposed in~\cite{song2020denoising} can be formulated as follows:
\begin{equation}
    \mathbf{x}_{t} = \sqrt{\bar{\alpha}_{t}}(\frac{\mathbf{x}_{t-1} - \sqrt{1-\bar{\alpha}_{t-1}}\epsilon_{\phi}(\mathbf{x}_{t}, t, \emptyset)}{\sqrt{\bar{\alpha}_{t-1}}}) + \sqrt{1 - \bar{\alpha}_{t} - \sigma^2_{t}}\epsilon_{\phi}(\mathbf{x}_{t}, t, \emptyset) + \sigma_{t}z^{*},
\end{equation}
where $\bar{\alpha} = \prod_{i=1}^{t} \alpha_{i}$, and $\alpha_{i}$ is the predefined constant of the timestep $i$, and $z^{*}$ is the white noise.
We use the $\mathbf{x}_{t-1}$ and $\mathbf{x}_{t}$ to denote the noisy latent, $\epsilon_{\phi}(\mathbf{x}_{t}, t, \emptyset)$ is modeled by the pretrained neural network $\phi$ with the unconditioned $\emptyset$.
When the hyperparameter $\sigma^2$ is set as $(1 - \alpha_{t})(1-\bar{\alpha}_{t-1})/(1 - \bar{\alpha}_{t})$, the DDIM is equivalent to DDPM~\cite{ho2020denoising}.
To accelerate the sampling process, DDIM set the $\sigma^2 = 0$, and the $\mathbf{x}_{t}$ can be approximately predicted from $\mathbf{x}_{t-1}$ as follows:

\begin{equation}
\begin{aligned}
    \mathbf{x}_{t} & = \sqrt{\bar{\alpha}_{t}}(\frac{\mathbf{x}_{t-1} - \sqrt{1-\bar{\alpha}_{t-1}}\epsilon_{\phi}(\mathbf{x}_{t}, t, \emptyset)}{\sqrt{\bar{\alpha}_{t-1}}}) + \sqrt{1 - \bar{\alpha}_{t}}\epsilon_{\phi}(\mathbf{x}_{t}, t, \emptyset) \\
    & \approx \sqrt{\bar{\alpha}_{t}} (\frac{\mathbf{x}_{t-1} - \sqrt{1 - \bar{\alpha}_{t-1}}\epsilon_{\phi}(\mathbf{x}_{t-1}, t, \emptyset)}{\sqrt{\bar{\alpha}_{t-1}}}) + \sqrt{1-\bar{\alpha}_{t}}\epsilon_{\phi}(\mathbf{x}_{t-1}, t, \emptyset).
\end{aligned}
\label{eq2}
\end{equation}    

Such approximation relies on the linear assumption $\epsilon_{\phi}(\mathbf{x}_{t}, t, \emptyset) \approx \epsilon_{\phi}(\mathbf{x}_{t-1}, t, \emptyset)$.
However, as indicated in~\cite{wallace2023edict}, this approximation results in the propagation of errors and thus leads to inconsistent loss in both forward and reverse processes.
\begin{algorithm}
    \caption{Exact Score Matching (ESM)}
    \begin{algorithmic}
        \STATE Initialization: DDIM inversion step size $\delta_{T}$ and $\delta_{S}$, the target prompt $y$, and the mixture ratio $\rho$
        \WHILE{$\Theta$ is not converged}
        \STATE Sample: $\mathbf{x}_{0} = \mathcal{R}(\Theta, C)$, $t \sim \mathcal{U}(1, 1000)$
        \STATE let $s = t - \delta_{T}, n = s/\delta_{S}$
        \FOR{i = [1, $\cdots$, {$n$}]}
        \STATE 
        $\mathbf{x}_{i\delta_{s}} = \sqrt{\bar{\alpha}_{i}}(\frac{\mathbf{x}_{(i-1)\delta_{s}} - \sqrt{1 - \bar{\alpha}_{i-1}}\epsilon_{\phi}(\mathbf{x}_{i\delta_{s}}, i\delta_{s}, \emptyset)}{\sqrt{\bar{\alpha}_{i-1}}}) + \sqrt{1 - \bar{\alpha}_{i}}\epsilon_{\phi}(\mathbf{x}_{i\delta_{s}}, i\delta_{s}, \emptyset)$
        
        \ENDFOR
        \STATE Copy a latent variable $\mathbf{x}'_{s}$ from $\mathbf{x}_{s}$, and parameterize $\epsilon_{lo}$ by a LoRA. 
        \STATE Let $\mathbf{x}_{s} \rightarrow \mathbf{x}_{int}$, $\mathbf{x}'_{s} \rightarrow \mathbf{x}'_{int}$ via
        \STATE $\mathbf{x}'_{int} = \sqrt{\bar{\alpha}_{t}}(\frac{\mathbf{x}'_{s}-\sqrt{1 - \bar{\alpha}_{t}}\epsilon_{lo}(\mathbf{x}_{s}, s, \emptyset)}{\sqrt{\bar{\alpha}_{s}}}) + \sqrt{1 - \bar{\alpha}_{s}}\epsilon_{lo}(\mathbf{x}_s, s, \emptyset)$
        \STATE $\mathbf{x}_{int} = \sqrt{\bar{\alpha}_{t}}(\frac{\mathbf{x}_{s}-\sqrt{1 - \bar{\alpha}_{t}}\epsilon_{lo}(\mathbf{x}'_{int}, s, \emptyset)}{\sqrt{\bar{\alpha}_{int}}}) + \sqrt{1 - \bar{\alpha}_{s}}\epsilon_{lo}(\mathbf{x}'_s, s, \emptyset)$
        \STATE Obtain the mixture noisy latent $\mathbf{x}_{t} = \frac{1}{\rho}[\mathbf{x}_{int} - (1 - \rho)\mathbf{x}_{int}']$
        \STATE Predict $\epsilon_{\phi}(\mathbf{x}_t, t, y)$ and $\epsilon_{\phi}(\mathbf{x}_s, s, \emptyset)$.
        \STATE 3D learnable representation $\Theta$ is updated by optimizing ESM loss
        \STATE $\nabla_{\Theta} \mathcal{L}_{\rm{ESM}} = \omega (t)(\epsilon_{\phi}(\mathbf{x}_t, t, y) - \epsilon_{\phi}(\mathbf{x}_s, s, \emptyset))$
        \ENDWHILE
    \end{algorithmic}
    \label{alg:esm}
\end{algorithm}
\subsubsection{Interval Score Matching (ISM)}
ISM is first introduced in Lucidreamer~\cite{liang2023luciddreamer} to alleviate the issue of the over-smoothing problem introduced by the Score Distillation Sampling (SDS)~\cite{poole2022dreamfusion}.
Specifically, for the given prompt $y$, the gradient of ISM loss is defined as:
\begin{equation}
    \nabla_{\Theta} \mathcal{L}_{\rm{ISM}}(\Theta) = [\omega(t)(\epsilon_{\phi}(\mathbf{x}_{t}, t, y) - \epsilon_{\phi}(\mathbf{x}_{s}, s, \emptyset))\frac{\partial \mathcal{R}(\Theta, C)}{\partial C}]\quad (0 < s < t),
\end{equation}
where $\mathcal{R}(\Theta, C)$ denotes the differential rendering function that is utilized to render an image of the camera pose $C$.
We use $\Theta$ to denote the learnable parameters of 3D representations, and $\omega(t)$ to represent a time-dependent function.
The noise latent $\mathbf{x}_{t}$ and $\mathbf{x}_{s}$ are calculated by the DDIM inversion process.
Given the DDIM inversion step size $\delta_{T}$, let $s = t - \delta_{T}$, and suppose $\mathbf{x}_{0}$ gradually approach to $\mathbf{x}_{s}$ in $n$ steps.
Hence, for each step $i$ $(1 \leq i \leq n)$, the forward diffusion process can be calculated via:
\begin{equation}
\begin{aligned}
\mathbf{x}_{i\delta_{s}} & = \sqrt{\bar{\alpha}_{i}}(\frac{\mathbf{x}_{(i-1)\delta_{s}}-\sqrt{1-\bar{\alpha}_{i-1}}\epsilon_{\phi}(\mathbf{x}_{i\delta_{s}}, i\delta_{s}, \emptyset)}{\sqrt{\bar{\alpha}_{i-1}}}) + \sqrt{1 - \bar{\alpha}_{i}}\epsilon_{\phi}(\mathbf{x}_{i\delta_{s}}, i\delta_{s}, \emptyset) \\
 & \approx \sqrt{\bar{\alpha}_{i}} (\frac{\mathbf{x}_{(i-1)\delta_{s}} - \sqrt{1 - \bar{\alpha}_{i-1}}\epsilon_{\phi}(\mathbf{x}_{(i-1)\delta_{s}}, i\delta_{s}, \emptyset)}{\sqrt{\bar{\alpha}_{i-1}}}) + \sqrt{1 - \bar{\alpha}_{i}}\epsilon_{\phi}(\mathbf{x}_{(i-1)\delta_{s}}, i\delta_{s}, \emptyset),
\end{aligned}
\end{equation}
where $\delta_{s}$ denote the step size in the approaching $\mathbf{x}_{0}\rightarrow \mathbf{x}_{s}$.
Then, $\mathbf{x}_{t}$ is calculated by $\mathbf{x}_{s}$ via:
\begin{equation}
\begin{aligned}
    \mathbf{x}_{t} & = \sqrt{\bar{\alpha}_{t}}(\frac{\mathbf{x}_{s} - \sqrt{1 - \bar{\alpha}_{s}}\epsilon_{\phi}(\mathbf{x}_{t}, t, \emptyset)}{\sqrt{\bar{\alpha}_{s}}}) + \sqrt{1 - \bar{\alpha}_{t}}\epsilon_{\phi}(\mathbf{x}_{t}, t, \emptyset) \\
    & \approx \sqrt{\bar{\alpha}_{t}}(\frac{\mathbf{x}_{s} - \sqrt{1 - \bar{\alpha}_{s}}\epsilon_{\phi}(\mathbf{x}_{s}, s, \emptyset)}{\sqrt{\bar{\alpha}_{s}}}) + \sqrt{1 - \bar{\alpha}_{t}}\epsilon_{\phi}(\mathbf{x}_{s}, s, \emptyset).
\end{aligned}
\end{equation}
Obviously, in the optimization of ISM, there are two approximations have been introduced which result in inconsistent loss during the diffusion processes and distortion in details.

\subsection{Exact Score Matching (ESM)}
\label{ESM}

\subsubsection{Motivation}
Since the necessary discrete computation of the underlying noise schedule, DDIM relies on the linear assumption to employ. 
To mitigate such approximation and enhance the consistent generation, inspired by~\cite{wallace2023edict}, we leverage the auxiliary variables to construct an exact recovery mathematically. 

Concretely, the $\mathbf{x}_{t-1}$ can be recovered from $\mathbf{x}_{t}$ via the approximation in Eq.~(\ref{eq2}).
Utilizing a new variable $\mathbf{x}_{t}'$ copied from $\mathbf{x}_{t}$, and then $\mathbf{x}_{t}$ can be exactly recovered via:
\begin{equation}
    \mathbf{x}_{t} = \sqrt{\bar{\alpha}_{t}}(\frac{\mathbf{x}_{t-1} - \sqrt{1 - \bar{\alpha}_t}\epsilon_{\phi}(\mathbf{x}_{t}', t,\emptyset)}{\sqrt{\bar{\alpha}_{t-1}}}) + \sqrt{1 - \bar{\alpha}_{t-1}}\epsilon_{\phi}(\mathbf{x}'_t, t, \emptyset).
\end{equation}
The same strategy can directly apply in ISM to mitigate the approximations that include two parts: $\mathbf{x}_{0} \rightarrow \mathbf{x}_{s}$ and $\mathbf{x}_{s} \rightarrow \mathbf{x}_{t}$.

\subsubsection{Optimization by ESM}

As analyzed above, we can mitigate the approximations by introducing extra independent variables.
However, extra variables will inevitably burden the computation, especially in multiple steps approaching $\mathbf{x}_{0} \rightarrow \mathbf{x}_{s}$. 
Considering this, we only mitigate the approximation in the one step approaching $\mathbf{x}_{s} \rightarrow \mathbf{x}_{t}$ in ISM.
Specifically, we first introduce $\mathbf{x}_{s}'$ that is copied from $\mathbf{x}_{s}$.
Then, noise intermediate latent $\mathbf{x}'_{int}$ and $\mathbf{x}_{int}$ can be calculated via:
\begin{equation}
\begin{aligned}
& \mathbf{x}'_{int} = \sqrt{\bar{\alpha}_{t}}(\frac{\mathbf{x}'_{s} - \sqrt{1 - \bar{\alpha}_{t}}\epsilon_{lo}(\mathbf{x}'_{s}, s, \emptyset)}{\sqrt{\bar{\alpha}_{s}}}) + \sqrt{1 - \bar{\alpha}_{s}}\epsilon_{lo}(\mathbf{x}_{s}', s, \emptyset), \\
& \mathbf{x}_{int} = \sqrt{\bar{\alpha}_{t}}(\frac{\mathbf{x}_{s} - \sqrt{1 - \bar{\alpha}_{t}}\epsilon_{lo}(\mathbf{x}_{int}', s, \emptyset)}{\sqrt{\bar{\alpha}_{s}}}) + \sqrt{1 - \bar{\alpha}_{s}}\epsilon_{lo}(\mathbf{x}_{int}', s, \emptyset).
\end{aligned}
\label{exa}
\end{equation}
Notably, the $\epsilon_{lo}$ is implemented by the LoRA~\cite{hu2021lora} of a pre-trained 2D diffusion model to effectively and adaptatively capture the noisy latent changing in the optimization process.
Moreover, to alleviate the diverging optimization of $\mathbf{x}_{int}$ and $\mathbf{x}_{int}'$, we interact them with a predefined ratio $\rho$ via $\mathbf{x}_{t} = \frac{1}{\rho}[\mathbf{x}_{int} - (1 - \rho)\mathbf{x}_{int}'], (0 < \rho \leq 1).$
Then, for the given prompt $y$, the 3D learnable representation $\Theta$ is optimized by minimizing $\mathcal{L}_{\rm{ESM}}$:

\begin{equation}
\begin{aligned}
    \mathcal{L}_{\rm{ESM}}(\Theta) = & \mathbb{E}_{t, C}[\omega(t)||\epsilon_{\phi}(\mathbf{x}_{t}, t, y) - \epsilon_{\phi}(\mathbf{x}_{s}, s, \emptyset)||^2]\\
     = & \mathbb{E}_{t, C} [\omega(t) (||\epsilon_{\phi}(\mathbf{x}_{t}, t, y) - \epsilon_{\phi}(\mathbf{x}_{int}, t, \emptyset)||^2  + ||\epsilon_{\phi}(\mathbf{x}_{int}, t, \emptyset) - \epsilon_{\phi}(\mathbf{x}_{s}, s, \emptyset)||^2)].
\end{aligned}
\end{equation}

Through introducing an auxiliary variable $\mathbf{x}_{s}'$ copied from $\mathbf{x}_{s}$, we can get an exact recovery from $\mathbf{x}_{s}$ to $\mathbf{x}_{t}$, and thus reduce the accumulate error occurs in DDIM inversion process.
The algorithmic pipeline of ESM is shown in Algorithm~\ref{alg:esm}.

\subsection{Comparison with ISM}
\label{com}
We now compare the differences between ISM and ESM mathematically.
Specifically, the optimization goal of ESM can be viewed to mitigate the error $\epsilon_{\rm{ISM}}$:
\begin{equation}
    \epsilon_{\rm{ISM}} = ||\epsilon_{\phi}(\mathbf{x}_{t}, t, y) - \epsilon_{\phi}(\mathbf{x}_{s}, s, \emptyset)||^2.
\end{equation}

In our ESM, as we introduce the auxiliary variable $\mathbf{x}_{s}'$ to maintain an exact recovery path, the optimization goal can be decoupled into two parts as follows:
\begin{equation}
    \epsilon_{\rm{ESM}} = ||\epsilon_{\phi}(\mathbf{x}_{t}, t, y) - \epsilon_{\phi}(\mathbf{x}_{int}, t, \emptyset)||^2 + ||\epsilon_{\phi}(\mathbf{x}_{int}, t, \emptyset) - \epsilon_{\phi}(\mathbf{x}_{s},s,\emptyset)||^2,
\end{equation}
where $\mathbf{x}_{int}$ is calculated via Equation~\ref{exa}.
Compared with ISM, ESM has a smaller accumulated error (\emph{i.e.}, $\epsilon_{\rm{ESM}} < \epsilon_{\rm{ISM}}$), and we show this conclusion in the follows: 

\begin{theorem}
    In diffusion-based 3D model generation, ESM is more effective than ISM in reducing accumulated error (i.e., $\epsilon_{\rm{ESM}} < \epsilon_{\rm{ISM}}$), thereby enhancing the detail fidelity of 3D representations. 
\end{theorem}
\textit{Proof:}
Let $\epsilon_{\phi}(\mathbf{x}_{int},t, \emptyset) = \epsilon_{\phi}(\mathbf{x}_s, s, \emptyset) + \eta$, and we assume $0 < \eta < ||\epsilon_{\phi}(\mathbf{x}_{t}, t, y) - \epsilon_{\phi}(\mathbf{x}_{s}, s, \emptyset)||$.
Then $\epsilon_{\rm{ESM}}$ can be represented as: 
\begin{equation}
\begin{aligned}
    \epsilon_{\rm{ESM}} & = ||\epsilon_{\phi}(\mathbf{x}_{t}, t, y) - (\epsilon_{\phi}(\mathbf{x}_{s}, s, \emptyset) + \eta)||^2 + ||(\epsilon_{\phi}(\mathbf{x}_s, s, \emptyset) + \eta) - \epsilon_{\phi}(\mathbf{x}_s, s, \emptyset)||^2 \\
    & = ||\epsilon_{\phi}(\mathbf{x}_{t}, t, y) - \epsilon_{\phi}(\mathbf{x_s}, s, \emptyset) - \eta||^2 + ||\eta||^2 \\
    & = ||\epsilon_{\phi}(\mathbf{x}_{t},t,y) - \epsilon_{\phi}(\mathbf{x}_{s}, s, \emptyset)||^2 - 2\eta || \epsilon_{\phi}(\mathbf{x}_{t}, t, y) - (\mathbf{x}_{s}, s, \emptyset)|| + 2 ||\eta||^2 \\
    & = \epsilon_{\rm{ISM}} + 2\eta(\eta - ||\epsilon_{\phi}(\mathbf{x}_{t}, t, y) - \epsilon_{\phi}(\mathbf{x}_{s}, s, \emptyset)||)
\end{aligned}
\end{equation}
Due to $2\eta (\eta - ||\epsilon_{\phi}(\mathbf{x}_{t}, t, y) - \epsilon_{\phi}(\mathbf{x}_{s}, s, \emptyset)||) < 0$, hence $\epsilon_{\rm{ESM}} < \epsilon_{\rm{ISM}}$.

\section{Experiments}
\subsection{Implementation details}
All experimental results are implemented by using a single NVIDIA 3090 RTX GPU.
The predefined parameters $\rho$, and $\{\delta_{S}, \delta_{T}\}$ are set as $0.93$, and $\{200, 50\}$ respectively.
We use the stable diffusion~\cite{rombach2022high} to implement the pre-trained diffusion model $\epsilon_{\phi}$, and follow~\cite{ryu2023low} to implement the LoRA.
Besides, we set the training process with 5,000 iterations and leverage the pre-trained Point-E~\cite{nichol2022point} to initialize the 3D gaussian splatting.

\subsection{Text-to-3D Generation}
The results generated by ExactDreamer are shown in Figure~\ref{fig:fig1}.
We input various types of prompts to illustrate the robustness of our method.
The results demonstrate that our method is capable of generating high-fidelity 3D models that consist of given texts, such as \textit{"a DSLR photo of a frog on a lotus leaf".}
Also, the photorealistic and imaginary 3D models can both be generated by ExactDreamer.

\subsection{Qualitative Comparison}
\begin{figure}[!t]
    \centering
    \includegraphics[width=1.0\linewidth]{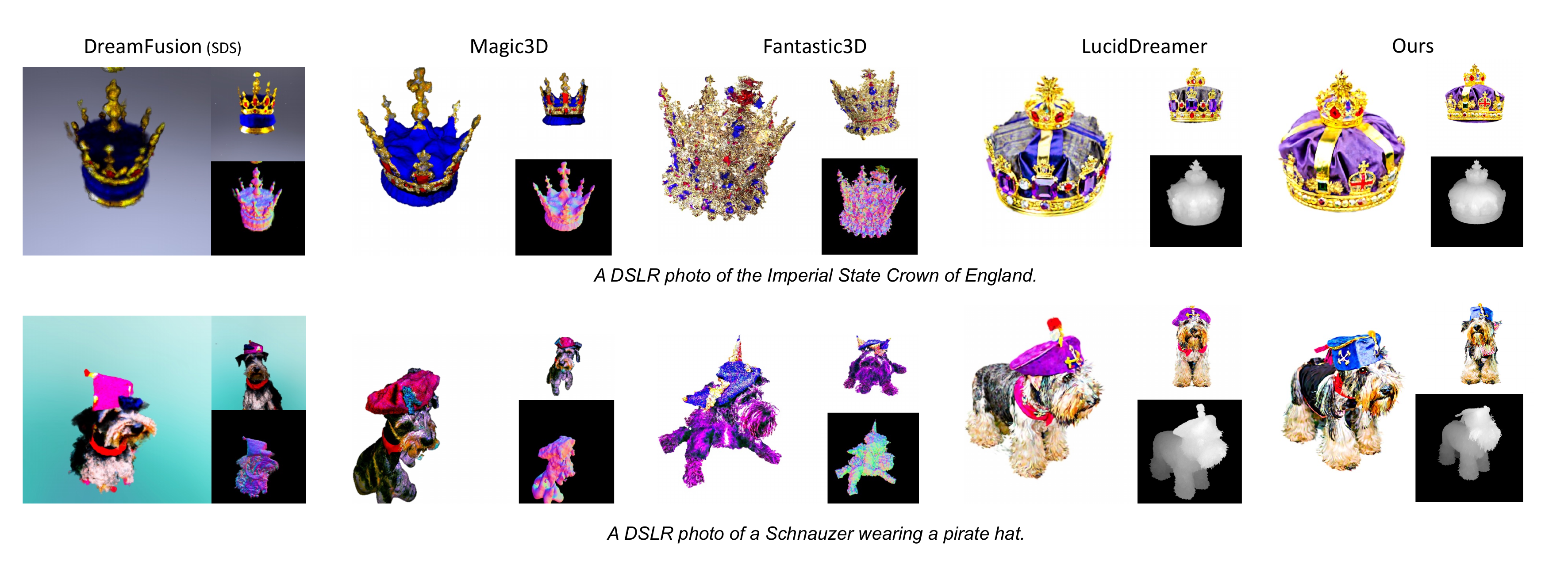}
    \caption{\textbf{Comparison with baselines in text-to-3D generation.} We compare our methods with current SoTA methods, and our method performs better in detail.}
    \label{fig:com}
\end{figure}

Utilizing the same prompt and setting in~\cite{liang2023luciddreamer}, we compare our method with current SoTA methods~\cite{poole2022dreamfusion,lin2023magic3d,chen2023fantasia3d,liang2023luciddreamer}.
The comparison results are shown in Figure~\ref{fig:com}. Our method, while maintaining high fidelity, is able to generate 3D models that are more consistent with the text compared to the baselines.
With the prompt \textit{"A DSLR photo of the Imperial State Crown of England"}, our method shows significant improvement in clarity compared to DreamFusion, Magic3D, and Fantasia3D. Compared to LucidDreamer, the crown we generated is more detailed and includes a national flag.
With the prompt \textit{"A DSLR photo of a Schnauzer wearing a pirate hat"}, our method excels in geometric shape and hair texture.
Compared to LucidDreamer, the Schnauzer we generated has more detailed eyes.

\subsection{Parameters Exploration}

\begin{figure}[!t]
    \centering
    \includegraphics[width=1.0\linewidth]{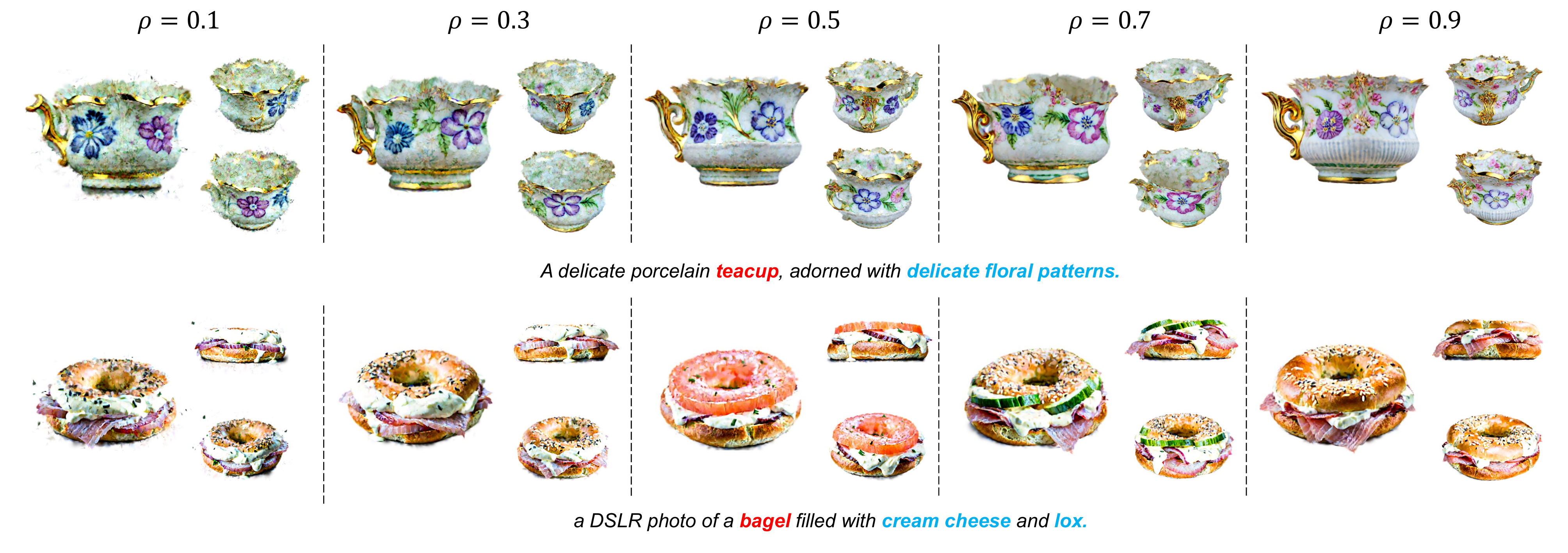}
    \caption{\textbf{Effect of mixed ratio.} We tune the $\rho$ in the interval $\{0.1, 0.3, 0.5, 0.7, 0.9\}$. 
    The generated results show that a higher mixed ratio is typically beneficial for high-fidelity generation. 
    }
    \label{fig:mix}
\end{figure}

\begin{figure}[!t]
    \centering
    \includegraphics[width=1.0\linewidth]{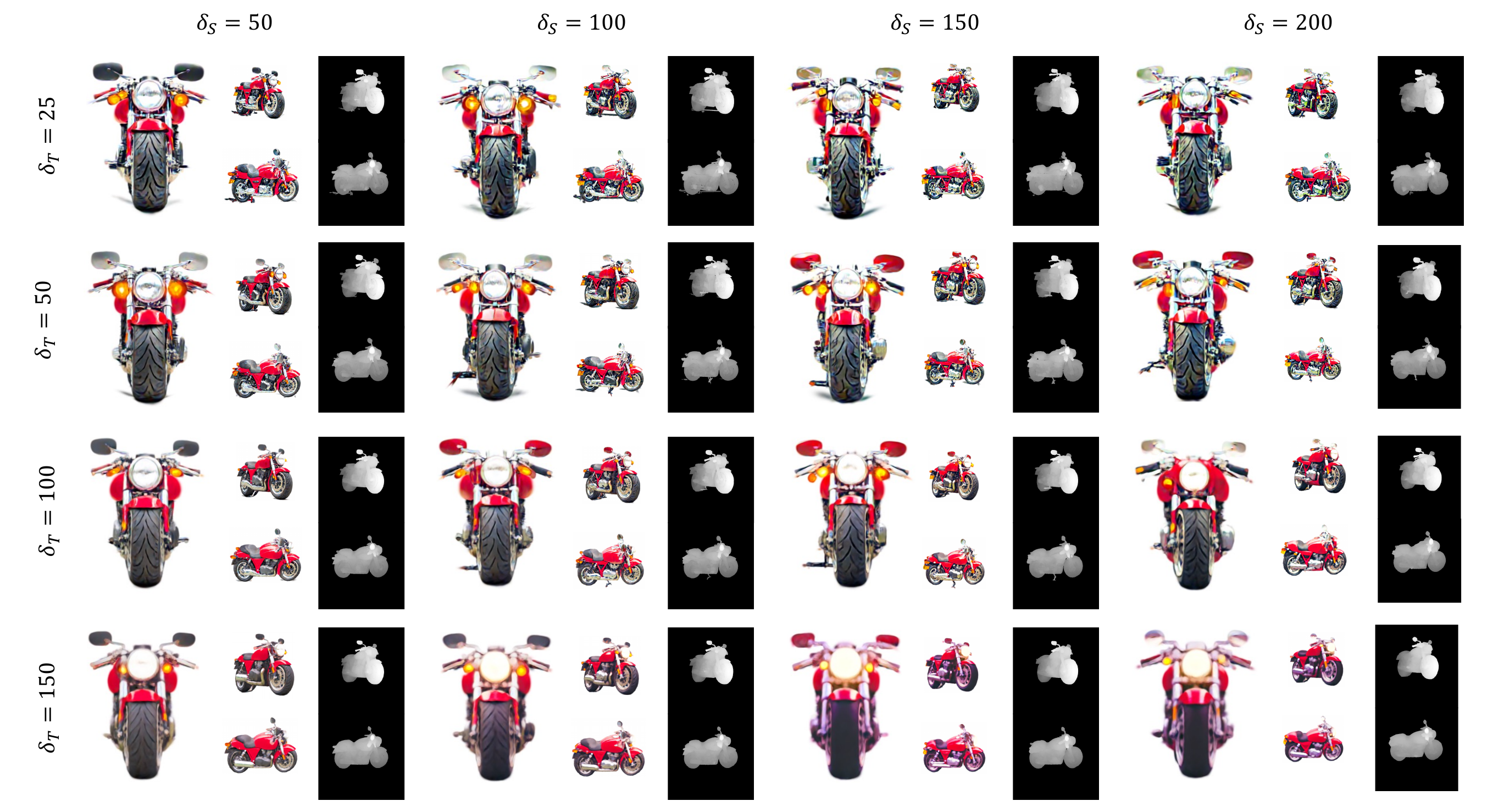}
    \caption{\textbf{Effect of step size.} The generated results are guided by the prompt \textit{A red motocycle}. The parameters $\delta_{S}$ and $\delta_{T}$ are tuned in the sets $\{50, 100, 150, 200\}$ and $\{25, 50, 150, 200\}$, respectively.
    The results illustrate that the step sizes have a significant influence on the clarity of generated 3D objections.
    Obviously, higher $\delta_{T}$ result in blurrier results.}
    \label{fig:delta}
\end{figure}
As shown in Algorithm~\ref{alg:esm}, the step sizes $\{\delta_{T}, \delta_{S}\}$ and mixture ratio $\rho$ are predefined hyperparameters.
In this section, we explore their effects under different settings on the final generation results.

\textbf{Mixed ratio $\rho$}~
The parameter $\rho$ quantifies the relative contributions of $\mathbf{x}_{int}'$ and $\mathbf{x}_{int}$.
We tune the $\rho$ within the set $\{0.1, 0.3, 0.5, 0.7, 0.9\}$ using two prompt guidance methods, and show the results in Figure~\ref{fig:mix}.
The results indicate that a higher  $\rho$ is beneficial for high-quality 3D model generation, whereas a lower $\rho$ may lead to unstable optimization. 

\textbf{Step sizes $\delta_{S}$ and $\delta_{T}$} With the prompt of \textit{A red motorcycle}, we tune the step sizes $\delta_{S}$ and $\delta_{T}$ within the sets $\{50, 100, 150, 200\}$ and $\{25, 50, 150, 200\}$ respectively, and present the results in Figure~\ref{fig:delta}.
The value of $\delta_{S}$ has no significant influence on the final results.
Therefore, we set $\delta_{S} = 200$ to conserve the computational resources in the DDIM inversion process.
While we mitigate the negative effects caused by the linear approximation, an increasing $\delta_{T}$ results an increasing accumulative error between $\mathbf{x}_{s}$ and $\mathbf{x}_{t}$.
This error leads to lower fidelity in the details of the generated motorcycle.

\subsection{Different Initialization}
\begin{figure}
    \centering
    \includegraphics[width=1.0\linewidth]{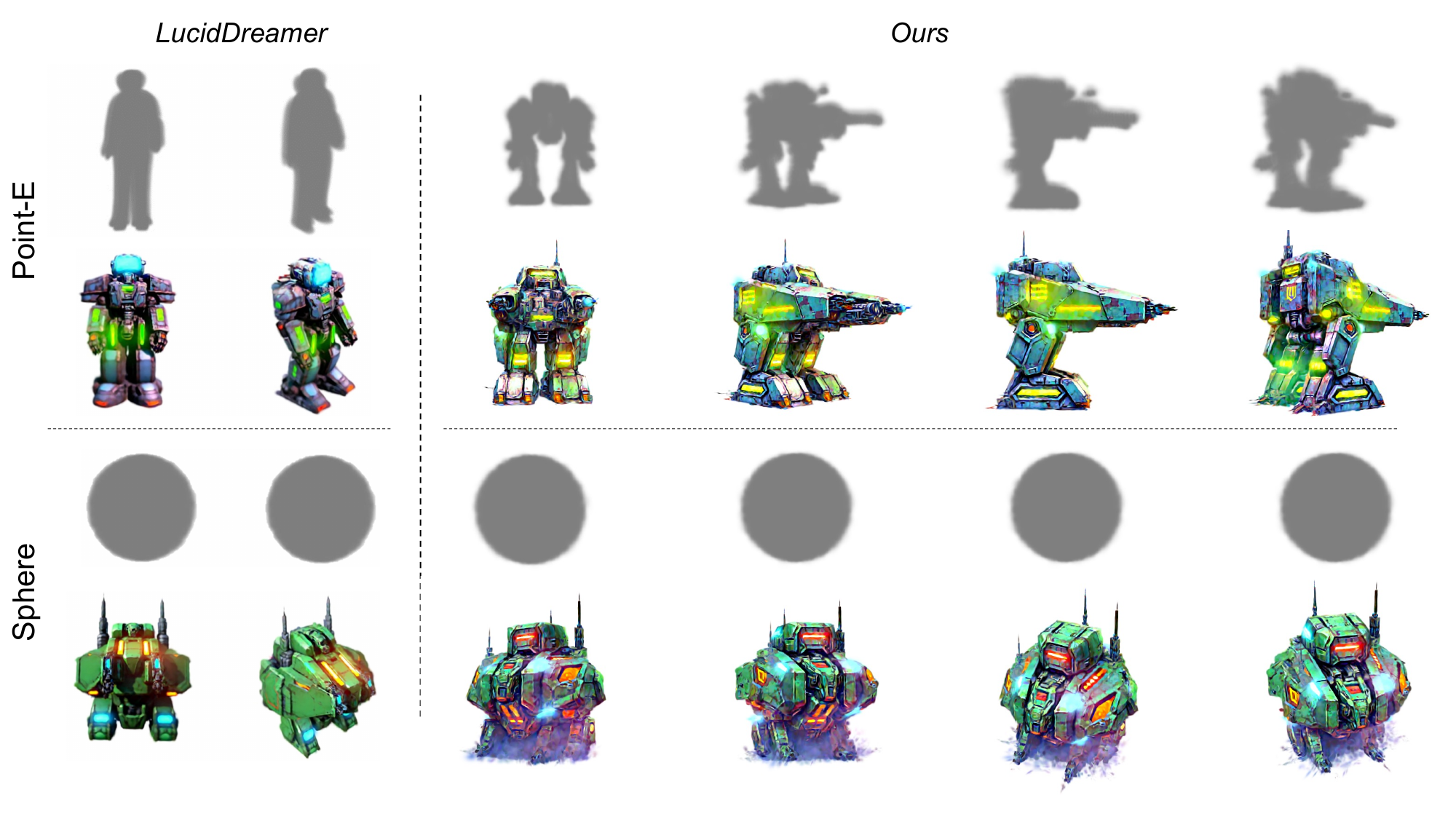}
    \caption{\textbf{Effect of different initialization.}
    Given the prompt \textit{"A military Mech, future, scifi"}, we compare the effect of two initialization, Point-E and Sphere, on the final generated results. 
    The results illustrate that Point-E provides better performance, producing more geometrically detailed models.
    Additionally, when compared with results generated by LucidDreamer, our method demonstrates a clear advantage in detail and overall quality.}
    
    \label{fig:initialization}
\end{figure}
Initialization has a significant impact on the final results. We compare two types of initialization: Point-E and Sphere, and present the results in Figure~\ref{fig:initialization}. 
Guided by the prompt \textit{"A military Mech, future, sci-fi"}, the geometric structure of the final generated results is largely determined by the initialization method.
Compared to Sphere, Point-E is able to generate more geometrically detailed results. 
We also compare the generated results with those from LucidDreamer. 
With Sphere initialization, our method is somewhat limited in generating the defined feet of the mech, but it produces more detailed features on the body of the military mech. 
However, with Point-E initialization, our method clearly outperforms LucidDreamer by capturing more complex details.
\section{Conclusion}

In this paper, we propose a novel approach named Exact Score Matching (ESM), based on Interval Score Matching (ISM), to enhance the consistency during diffusion-based 3D generation processes. 
Specifically, we first analyze the inherent bias caused by the linear approximation adopted in DDIM, which is unstable and leads to inconsistent generation. 
To address this issue, we introduce an auxiliary variable and construct an exact recovery algorithm that can alleviate the accumulative error caused by the linear assumption in ISM. 
Additionally, to effectively and adaptively capture the dynamic changes of the original and auxiliary variables, we leverage LoRA to implement the recovery path. 
Our overall approach, named ExactDreamer, can be implemented using a single NVIDIA 3090 RTX GPU, and the results demonstrate that ExactDreamer can generate high-fidelity 3D content. However, our methodology does exhibit a few minor limitations, including unstable generation and sensitivity to hyper-parameters, we plan to polish these issues in future improvements. It's also important to acknowledge that while our research is directed towards improving the quality of generative models, it may help in the development of deepfake technology.

\bibliographystyle{plain}
\bibliography{ref}

\clearpage


\end{document}